\documentclass[runningheads]{llncs}

\usepackage{eccv}

\usepackage{eccvabbrv}
\usepackage{pgfplots}

\usepackage{graphicx}
\usepackage{booktabs}

\usepackage[accsupp]{axessibility}  
\usepackage{multicol}
\usepackage{multirow}
\usepackage{lipsum}
\usepackage{array}

\usepackage{hyperref}

\usepackage{orcidlink}

\usepackage{comment}
\usepackage{bibentry}
\nobibliography*
\excludecomment{suppcontent} 

\begin{document}


\title{\texorpdfstring{Missing Modality Prediction\\for Unpaired Multimodal Learning\\ via Joint Embedding of Unimodal Models}{Missing Modality Prediction for Unpaired Multimodal Learning  via Joint Embedding of Unimodal Models}}


\titlerunning{Missing Modality Prediction for Unpaired Multimodal Learning}

\author{Donggeun Kim\orcidlink{0009-0000-0900-0099} \and Taesup Kim\orcidlink{0009-0005-6056-6836}\thanks{A corresponding author}}

\authorrunning{D. Kim and T. Kim}

\institute{Graduate School of Data Science, Seoul National University \\
\email{\{kdg5188, taesup.kim\}@snu.ac.kr}}

\maketitle
\begin{abstract}
Multimodal learning typically relies on the assumption that all modalities are fully available during both the training and inference phases. However, in real-world scenarios, consistently acquiring complete multimodal data presents significant challenges due to various factors. This often leads to the issue of missing modalities, where data for certain modalities are absent, posing considerable obstacles not only for the availability of multimodal pretrained models but also for their fine-tuning and the preservation of robustness in downstream tasks. To address these challenges, we propose a novel framework integrating parameter-efficient fine-tuning of unimodal pretrained models with a self-supervised joint-embedding learning method. This framework enables the model to predict the embedding of a missing modality in the representation space during inference. Our method effectively predicts the missing embedding through prompt tuning, leveraging information from available modalities. We evaluate our approach on several multimodal benchmark datasets and demonstrate its effectiveness and robustness across various scenarios of missing modalities.

\end{abstract}
\section{Introduction}
Humans perceive the world through various sensory modalities, such as seeing images and hearing voices, integrating these diverse sources to enhance comprehension. Similarly, a fundamental goal in artificial intelligence is to equip computers with the capability to effectively learn from multi-sensory data. Multimodal learning emerges as a promising approach to improve our understanding of complex data by leveraging multiple communicative modalities.
In this context, significant advancements have been made in multimodal learning, particularly through self-supervised learning within the vision-language domain \cite{sun2019videobert, tan2019lxmert, chen2020uniter, clip}.
Furthermore, remarkable progress has been achieved in audio-visual learning \cite{audio-visual}, as well as in other multiple modalities.

However, in real-world scenarios, collecting completely paired data presents challenges for various reasons, such as data privacy and security issues. 
Consequently, the assumption that multimodal learning necessitates the completeness of all modalities is challenging to maintain, leading to the unpaired data issue and the resulting problem of missing modalities. Prior works \cite{Ma_2022_CVPR,Lee_2023_CVPR} employed multimodal pretrained models for handling missing modality in the fine-tuning stage. But missing modality due to unpaired data can also occur in the pretraining stage. For instance, in the medical domain, most medical datasets typically consist of images, the majority of which lack accompanying text-based clinical narrative reports. Even though it has shown potential when training with paired data \cite{medical2,medical1}, a significant portion of existing medical datasets, which include either image-only or text-only data, remains underutilized. As a result, there may be scenarios where obtaining an effective joint (multimodal) encoder, pretrained on hundreds of millions of image-text pairs, becomes challenging.

On the other hand, acquiring unimodal data is relatively easier than obtaining multimodal data. 
Additionally, pretrained unimodal models, which come in various forms and demonstrate high performance, are more readily accessible than multimodal models.
Hence, in this paper, we acknowledge the commonality of scenarios involving only unpaired data and consider that a pretrained joint encoder model in previous approaches \cite{Ma_2022_CVPR,Lee_2023_CVPR} may not always be available. 
Instead, we propose utilizing independently pretrained unimodal encoders for each modality. 
This strategy offers relatively broader applicability, as each unimodal encoder can be effectively trained using self-supervised learning with large-scale unlabeled data. \cite{SimCLR, BYOL, swav,SimSiam}.
Moreover, it benefits from leveraging knowledge gained during pretraining. For example, most multimodal models in the vision-language domain have primarily focused on datasets containing images and English-based text.
However, these approaches may encounter challenges in handling low-resource or multilingual languages \cite{multilingaulwiki,Multilingualfashion}. This issue can be easily addressed by using a (domain-specific) text encoder, which is pretrained on large, unlabeled datasets in those languages.

In this paper, we define the problem for multimodal settings as follows: (1) Pretrained unimodal encoders are assumed to exist; (2) Partially unpaired data for downstream tasks is provided; (3) Unpaired data is also given during inference. Our assumption is realistic and directly applicable to the multimodal downstream task in real-world scenarios. To this end, we propose a straightforward yet effective framework that addresses missing modalities by leveraging unimodal pretrained encoders and predicting the representations of the missing modalities.
We achieve this by our contributions, summarized as follows:
\begin{itemize}
    \item We utilize Parameter-Efficient Fine-Tuning (PEFT) to minimally update pretrained unimodal encoders while maximally preserving knowledge for downstream tasks.
    \item We employ the architecture of Variance-Invariance-Covariance Regularization (VICReg) for improving the predictability of embeddings between different modalities for missing modality problem.
    \item We adopt a prompt-based approach for gathering efficient task-relevant information from other modalities.
    \item We demonstrate that our approach is more robust and effective, outperforming previous studies across all tested datasets and metrics in various scenarios with missing modalities.
\end{itemize}

\section{Related Work}

\subsubsection{Missing Modality in Multimodal Learning}
Specific modalities available during training may be unavailable at inference, posing challenges in multimodal learning. \cite{Ma_2022_CVPR} investigated the robustness of pretrained multimodal transformers when encountering incomplete modalities during inference. For addressing a more general scenario where the absence of modality may occur during either the training or testing phase, \cite{Lee_2023_CVPR} leveraged missing-aware-prompts according to the missing case. However, both works rely on multimodal joint encoder pretrained with extensive image-text pairs, assuming the availability of large paired multimodal datasets. This assumption limits applicability in scenarios lacking such paired data. In contrast, our method uses unimodal models trained on unpaired data, respectively, eliminating the need for extensive paired datasets. Therefore, we offer a more adaptable approach to handling missing modalities, enhancing flexibility in multimodal learning without reliance on paired data.

There has also been some progress in addressing missing modalities by inferring them by modeling the probabilistic relationships between modalities. \cite{ma2021smil} proposes a method of Bayesian Meta-Learning to estimate the latent feature of the modality-incomplete data and reconstruct the features of the missing modality data. \cite{wang2023multi} proposes a strategy using shared encoder features from available modalities to generate modality-specific features of missed modality. These approaches are similar to ours in utilizing modality-specific encoders; however, our method focuses explicitly on the effective use of unimodal pretrained models. By efficiently fine-tuning unimodal models—widely available and pretrained on extensive unlabeled datasets—we ensure the maximal preservation of knowledge from pretraining. This distinction underscores our method's versatility and adaptability, suggesting its potential effectiveness in low-resource scenarios, such as handling multilingual text-image data.


\subsubsection{Parameter-Efficient Transfer Learning}

As the field advances with substantial pretrained models based on transformer\cite{vaswani2017attention} architecture, various parameter-efficient adaptation approaches \cite{zaken2021bitfit,houlsby2019parameter,hu2021lora} have emerged, approximating the performance of full fine-tuning by updating only a subset of parameters. Concurrently, prompt-based learning  \cite{li2021prefix, prompt}, initially successful in natural language processing, has shown promising results in computer vision tasks \cite{vpt, l2p,wang2022dualprompt, sprompt} as well, notably with vision transformer \cite{vit}. Inspired by these approaches, many recent works \cite{coop,cocoop,khattak2023maple,promptsrc} have been explored for adapting large pretrained vision-language models (\eg, CLIP \cite{clip}) without re-training the entire model. These methods offer the advantage of maintaining the knowledge learned from a large multimodal dataset while efficiently adapting to the target task.

In the pursuit of efficient training for multimodal learning with unimodal encoders, \cite{khan2023contrastive} demonstrated a method for parameter-efficient vision-language alignment by leveraging pretrained unimodal models. It achieves significant alignment with minimal image-text pairs and parameter updates, thus preserving the existing knowledge within pretrained models. Inspired by this approach, our framework utilizes separate unimodal pretrained encoders. This strategy provides flexibility for various input modalities and enhances efficiency by limiting updates to a subset of parameters across the entire model.

\subsubsection{Joint Embedding Predictive Architecture}
The Joint Embedding Predictive Architecture (JEPA) \cite{lecun2022path} combines embedding modules with latent variables and supports dual encoders generating distinct representations without sharing parameters. Its flexibility allows for processing various data formats, such as multimodal inputs. The primary training objective of JEPA is to establish predictability between these representations in the embedding space, and it competes against traditional contrastive methods that require a considerable number of negative samples or memory banks. Our framework aligns with JEPA's principles, focusing on developing encoders trained to infer the representation of one modality from another within the paired multimodal dataset.

We combine JEPA in a multimodal learning setup with Variance-Invariance-Covariance Regularization(VICReg) \cite{bardes2021vicreg}. VICReg was introduced to prevent the occurrence of \textit{collapse} in embeddings, where encoders produce constant or uninformative vectors while producing content features that transfer well on many downstream tasks. We adopt the VICReg objective for training our framework, which shares architectural similarities with JEPA, due to its simplicity both mathematically and computationally. Drawing from ideas described in \cite{bardes2021vicreg}, we adapt the objective to learn predictability from paired multimodal data. In this work, VICReg plays a crucial role in facilitating the prediction of features in the representation space, aligning with our objective of maximizing mutual information between the embedding of missing modality and the prediction \cite{shwartz2023information}. 

\section{Preliminaries}
\subsection{Problem Definition}

To maintain simplicity without losing generality, we consider a multimodal problem setting consisting of two modalities ($M=2$), namely $m_1$ and $m_2$ (e.g., image and text).
We further assume that these modalities do not always coexist, indicating \textit{the presence of missing modalities throughout both training and testing phases}.
Therefore, given a multimodal dataset $D=D^c \cup D^{m_1} \cup D^{m_2}$, it can be divided into three subsets: the modality-complete subset $D^c=\left\{\left(x_i^{m_1}, x_i^{m_2}, y_i\right)\right\}$ and two modality-incomplete subsets $D^{m_1}=\left\{\left(x_j^{m_1}, y_j\right)\right\}$ and $D^{m_2}=\left\{\left(x_k^{m_2}, y_k\right)\right\}$ (e.g., image-only and text-only).

Building on these assumptions, we additionally posit that there is no pretrained multimodal encoder available for processing multimodal data. 
Instead, each modality is supported by its own pretrained unimodal encoder, which has been trained independently without awareness of the other modality.
For this reason, we focus on more general problem settings that can be easily applied when only pretrained unimodal encoders are available.

\begin{figure*}[tp]
    \centering
    \includegraphics[width=\textwidth]{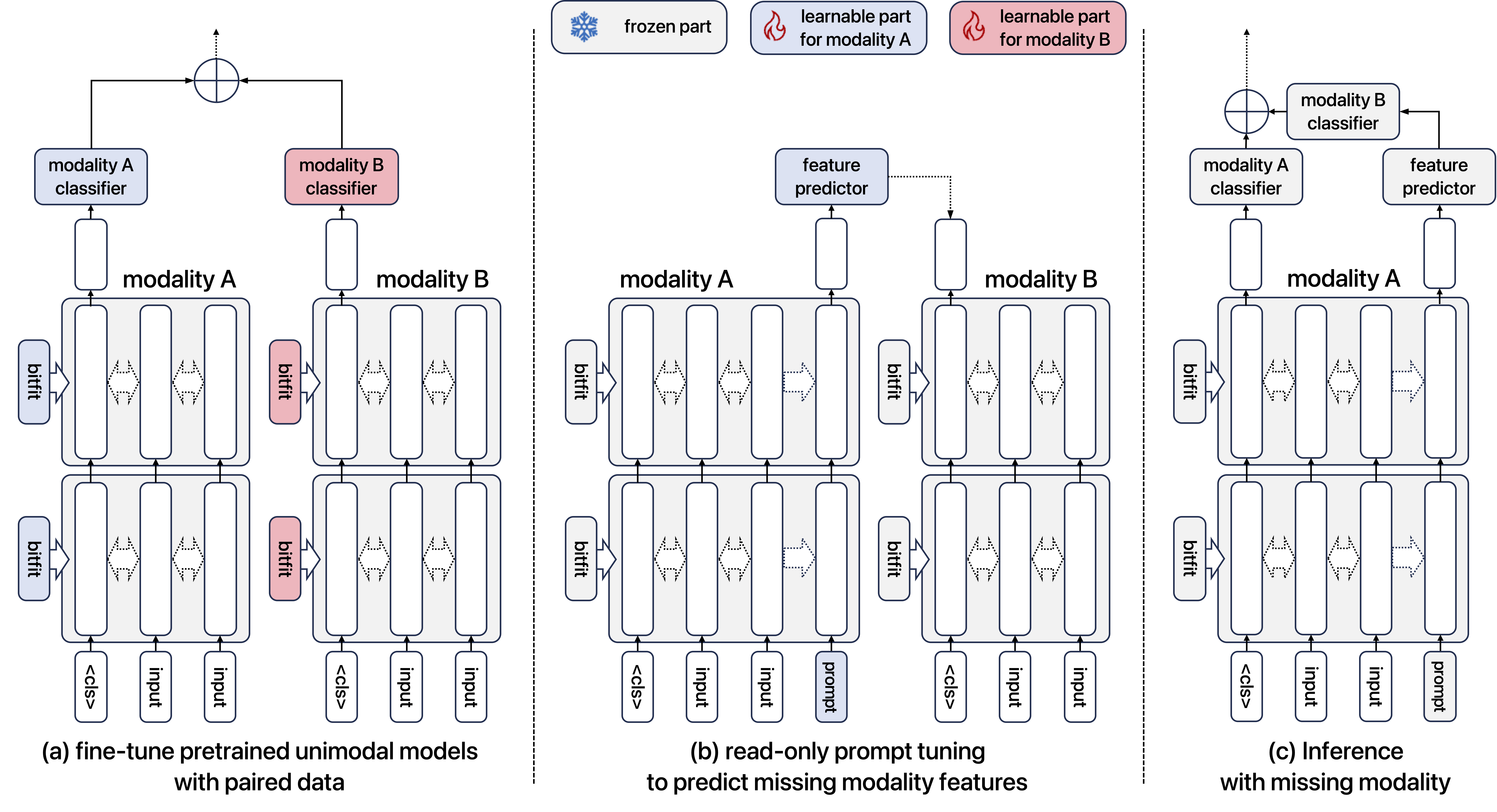}
    \caption{\textbf{Our framework overview}: Utilizing separate unimodal encoders, we employ PEFT for target tasks and introduce trainable prompts for effectively predicting missing modality features, with attention masking to preserve input's representations. During inference, model can generate embeddings of missing modality at hand.}
    \label{fig:overview}
\end{figure*}

As shown in Fig. \ref{fig:overview}.(a) and (b), for each modalities $m\in \{m_1, m_2\}$, we assume that a pretrained unimodal encoder $f_{\theta_{\text{enc}}^{m}}$ based on transformer \cite{vaswani2017attention} architecture is given and a classifier $f_{\theta_{\text{cls}}^{m}}$ is defined on top of its representation. 
We implement a straightforward late-fusion strategy, integrating the pre-softmax logits from each modality. 
To address the challenge posed by missing modalities, we introduce a feature predictor $f_{\theta_{\text{prd}}^{m}}$, designed to predict the feature vector of a missing modality.
Furthermore, to enhance its prediction capabilities, we employ a set of trainable prompts $\phi^{m}$.
Based on this setting, we aim to construct a multimodal model against challenges arising from incomplete multimodal data issues during both training and testing scenarios.

\subsection{Read-only Prompts}\label{prompts_section}
To enhance the capability of predicting embeddings of other modalities, it is necessary to update parameters of encoders or train only the newly added parameters. However, such updates potentially compromise the high-quality representations obtained through unimodal pretraining. Prompt tuning is an effective strategy to mitigate this, which adds learnable tokens to the input sequence without altering the encoder's pretrained weights. This method can facilitate model adaptation while preserving the original weights. Nevertheless, conventional prompt tuning can still affect the original representation due to the attention interaction between input data and prompts. To address this, we employ a read-only prompting technique \cite{lee2023read}. By applying a masking strategy to the self-attention mechanism specifically targeting the interaction between input data and prompt tokens, we ensure that only the prompts can "read" the input token features. This approach keeps token features unaffected by the prompts, allowing the prompts to focus on extracting relevant information necessary for feature prediction across modalities. Consequently, the prompts become specialized for the sub-task, feature prediction.

\section{Proposed Method}

\subsection{Multimodal Classification Task through a Simple Late-Fusion Strategy}\label{sec:cls}
Although pretrained unimodal encoders with a late-fusion strategy generally perform well without fine-tuning, they may not be sufficient to attain optimal performance in certain multimodal downstream tasks. However, full fine-tuning, while potentially improving performance, is less favorable due to its significant memory and resource demands. Therefore, we employ BitFit \cite{zaken2021bitfit} as a PEFT approach, which freezes all parameters of the entire model and updates only the bias terms during fine-tuning.
Based on this setting, we define the multimodal classification loss $L_{\text{cls}}$ as the summation of standard cross-entropy classification losses over multiple modalities as follows:
\begin{equation*}
L_{\text{cls}}
=
L_{m_1}\left(D^{m_1}; \theta_{\text{enc}}^{m_1}, \theta_{\text{cls}}^{m_1}\right)
+
L_{m_2}\left(D^{m_2}; \theta_{\text{enc}}^{m_2}, \theta_{\text{cls}}^{m_2}\right)
+
L_{c}\left(D^c; \theta_{\text{enc}}^{m_1},\theta_{\text{enc}}^{m_2}, \theta_{\text{cls}}^{m_1},\theta_{\text{cls}}^{m_2}\right)
\end{equation*}
where $\mathcal{L}_{m_1}$, $\mathcal{L}_{m_2}$ are the loss for modality-incomplete subsets, and $\mathcal{L}_{c}$ is the loss for a modality-complete subset.
As our framework is based on a late-fusion strategy, it enables our approach to be compatible with any other PEFT methods, such as adapter-based tuning \cite{houlsby2019parameter} or reparametrization-based method \cite{hu2021lora}.

\subsection{Missing Modality Feature Prediction with Prompt-Tuning}
In the presence of a missing modality, instead of using only the features of the existing modalities, we posit that the predicted features of the missing modality can be integrated with those of the available modalities during inference to enhance prediction performance. Consequently, we introduce a feature predictor $f_{\theta_{\text{prd}}^m}$ using a set of trainable prompts $\phi^m$ to address the issue of missing modalities effectively.
To facilitate this, we utilize \textit{read-only prompts} described in \ref{prompts_section} that are concatenated to the unimodal input data and then processed through an unimodal encoder based on transformers with specially designed masked attention.
This makes our feature predictor only read the internal representation of the encoder $f_{\theta_{\text{enc}}^m}$, which is fine-tuned for the downstream task, and to learn to utilize rather than modify it.
More precisely, we define the input data for each modality $m$ as \smash{$\mathbf{x}^{m}=[C^{m}, E^{m}, \phi^m]$} and the corresponding outputs can be expressed as:
$$
f_{\theta_{\text{enc}}^{m}}(\mathbf{x}^{m}) = \left[\widetilde{C}^{m}, \widetilde{E}^{m}, \widetilde{\phi}^m\right]
$$
where both the class token(\ie, CLS) embedding $\widetilde{C}^m$ and the input token embeddings $\widetilde{E}^m$ remain unchanged regardless of the prompts $\phi^m$ due to the use of read-only prompts.
Based on these output embeddings, the class prediction $f_{\theta_{\text{cls}}^m}(\widetilde{C}^{m})$ for the existing modality $m$ is computed, and the feature (\ie, the final embedding of a class token) of the absent modality $m'$ is predicted as (also see Fig. \ref{fig:overview}-(b)) $\hat{C}^{m'}=f_{\theta_{\text{pred}}^{m}}(\widetilde{\phi}^m)$.
It is important to note that the class prediction is not entirely affected by the feature prediction.
Furthermore, the feature prediction can be enhanced by only tuning the prompts without interfering with the internal representation in the unimodal encoders.

To optimize our feature prediction, we utilize a modality-complete dataset $D^c$ and simulate the situations of missing modality.
Moreover, to improve the predictability of embeddings similar to the approach outlined in \cite{lecun2022path}, we adopt VICReg \cite{bardes2021vicreg}. 
The loss function based on it for predicting embeddings while preventing their collapse comprises three components. 
Firstly, a variance term forces the embedding vectors of samples within a batch to be different. 
It involves a hinge loss function that maintains the standard deviation of each component of the embeddings along the batch dimension. Secondly, an invariance term is the main objective, mean-squared euclidean distance computed between the original and predicted features.
Finally, a covariance term is incorporated to decorrelate the different dimensions of the embeddings by setting the off-diagonal coefficients in the covariance matrix of the embeddings to zero.
Therefore, the loss function for our feature prediction $L_{\text{prd}}$ is a weighted average of the invariance, variance, and covariance terms:
\begin{equation}\resizebox{.95\textwidth}{!}{$
L_{\text{prd}}\left(\widetilde{C}^{m'},\hat{C}^{m'};\theta_{\text{prd}}^{m},\phi^m\right)
=
\lambda s\left(\widetilde{C}^{m'},\hat{C}^{m'}\right)+\mu\left[v(\widetilde{C}^{m'})+v(\hat{C}^{m'})\right]+\nu\left[c(\widetilde{C}^{m'})+c(\hat{C}^{m'})\right]$}
\label{eq:vicreg}
\end{equation}
where $s$, $v$ and $c$ are the invariance, variance and covariance terms as described in \cite{bardes2021vicreg} and $\lambda, \mu$ and $\nu$ are hyper-parameters.
Moreover, this loss function is based on the existing modality $m$ and the missing modality $m'$, and it can be applied vice versa with a modality-complete dataset $D^c$.
For this reason, our method can efficiently address any type of missing modality scenario.
Furthermore, to guide the feature predictor in generating features suitable for the downstream task and enhance the robustness of the classifier in missing cases, we introduce an auxiliary classification loss, $L_{\text{aux}}$. 
This can be achieved by taking the predicted features $\hat{C}^{m'}$ to the classifier and optimizing them with cross-entropy loss. It simulates the situations of missing modality and ensures that the predicted representation aligns effectively with the downstream task.
To sum up, the overall objective function $L_{\text{total}}$ can be represented as follows:
\begin{equation*}
L_{\text{total}}=\alpha *(L_{\text{cls}}+L_{\text{aux}})+L_{\text{prd}}
\end{equation*}
where $\alpha$ is the balancing hyper-parameter.

\section{Experiment}
\subsubsection{Dataset}
We evaluate our proposed method using three multimodal classification datasets following prior works \cite{Ma_2022_CVPR,Lee_2023_CVPR}. MM-IMDb \cite{arevalo2017gated} consists of 25,956 image-text pairs with movie plot outlines and poster images. This encompasses 23 different genre classes, and the objective is to predict the genres of movies. As movies are frequently associated with multiple genres, the task is multimodal multi-label classification. UPMC Food-101 \cite{wang2015recipe} is a multimodal classification dataset that includes images obtained by Google Image Search and corresponding textual descriptions for 101 food types. Comprising 90,840 image-text pairs, the dataset captures real-world challenges due to the noisy nature of the image and text pairs. Hateful Memes \cite{kiela2020hateful} is a challenging dataset designed to identify hate speech in memes using both image and text modalities. The selection of memes is structured to challenge strictly unimodal classifiers, making them struggle to classify correctly, while multimodal models are more likely to perform better. Hateful Memes emphasizes the importance of multimodal approaches in mitigating the limitations of unimodal signals. 
\subsubsection{Metric}
For MM-IMDb, we measure multi-label classification performance using the F1-Macro score; for UPMC Food-101, we compute the classification accuracy; and for Hateful Memes, we assess performance using the Area Under the Receiver Operating Characteristic Curve (AUROC).

\subsection{Experiment Setting} 
This paper explores two training settings involving missing modalities: complete training setting and missing training setting. Throughout our experiments, we compare our method with previous works \cite{Ma_2022_CVPR,Lee_2023_CVPR} based on a multimodal encoder (ViLT \cite{kim2021vilt}) and an unimodal baseline. The unimodal baseline employs encoders identical to ours but only leverages an unimodal classifier from the available data when a modality drops. It should be noted that our framework is adaptable to any training setting, while prior works lack the flexibility to apply to both settings effectively. This highlights the distinct advantage of our approach, which can handle various missing scenarios. 

\textit{Complete training setting} involves training on modality-complete data $D^c$, and evaluating on modality-incomplete data $D^{m_1}$(\ie, image-only (text-missing) data). This setup is designed to measure the model's robustness in the absence of the dominant modality. We compare our method with a previous work \cite{Ma_2022_CVPR} with a multimodal pretrained transformer. We have replicated the results from \cite{Ma_2022_CVPR} as no official code is available. Additionally, we compare unimodal methods against a vision-only encoder trained and tested solely on image modality data to measure the missing modality's robustness. For a fair comparison, results
 are averaged over five different random seeds, thereby enhancing the validity of the results by accounting for variability in the absence of text during testing.

\textit{Missing training setting} is a more general and challenging scenario where modality is absent in both the training and testing phases. We set a 70\% missing rate in all our experiments following \cite{Lee_2023_CVPR}. We explore three realistic cases of missing modality: text-missing, image-missing, and both-missing scenarios, and evaluate each case for all three scenarios of missing modality. To compare with previous work \cite{Lee_2023_CVPR} under our experimental setting, we reproduced it using the official code\footnote{\url{https://github.com/YiLunLee/missing_aware_prompts/tree/main}}. This was necessary as the earlier study only conducted experiments under the condition that the missing setting in training was equal to the testing phase. The performance metrics are calculated by averaging outcomes across five different random seeds. The seed determines which samples contain missing modality and which modality is absent, ensuring a fair comparison. 
\subsection{Main Result} 
\begin{figure*}[tp]
    \centering
    \includegraphics[width=\textwidth]{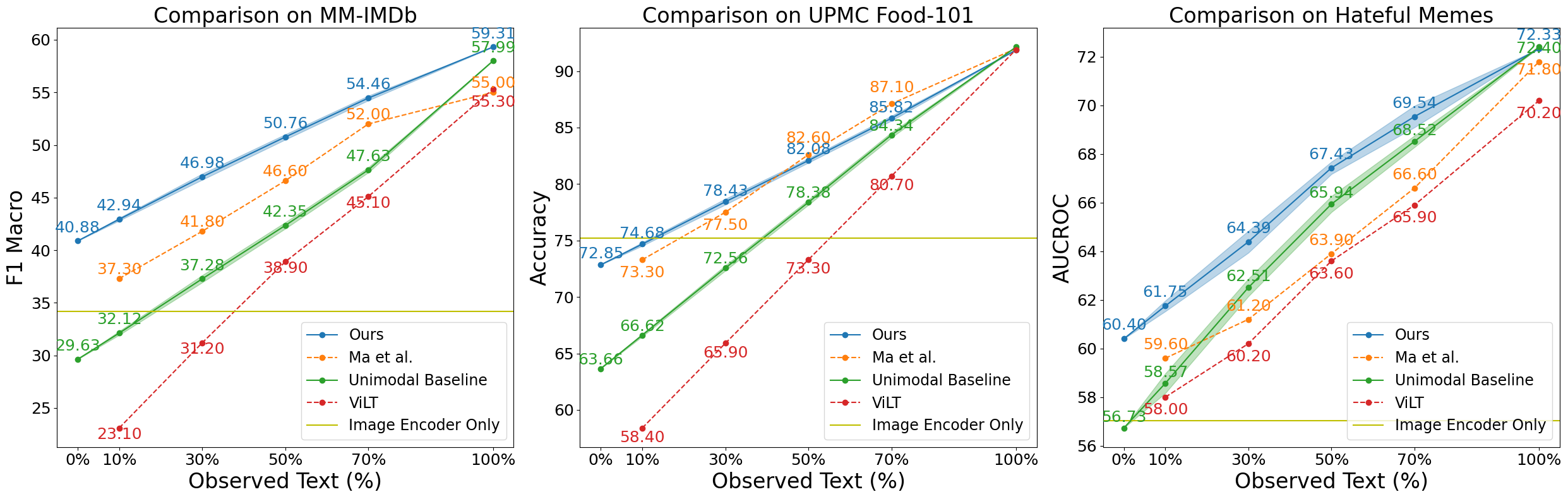}
    \caption{\textbf{Performance on multimodal classification datasets under a complete training setting}. All experiments were conducted with 100\% text and 100\% image data and evaluated based on the text missing rate. Since the official code from \cite{Ma_2022_CVPR} is unavailable, we have referenced the results for the multimodal model from prior work \cite{Ma_2022_CVPR}. Unimodal-based methods were evaluated by averaging performances obtained with five different seeds. Dotted lines indicate multimodal models, whereas solid lines represent unimodal models.}
    \label{fig:result1}
\end{figure*}
\begin{table*}[tp]
\centering
\caption{
\textbf{Quantitative results obtained under a training setting with a 70\% missing rate}. 
Using official code, we reproduced the results of previous works that utilize a multimodal encoder, specifically ViLT \cite{kim2021vilt} and \cite{Lee_2023_CVPR}. Evaluations were conducted by averaging the results from five different seeds across various modality-missing scenarios.}
\label{missing_result}
\resizebox{\textwidth}{!}{
\begin{tabular}{c|cc|cc|c|c|c|c} 
\toprule
\multirow{2}{*}{Dataset}                                                         & \multicolumn{2}{c|}{Training}                   & \multicolumn{2}{c|}{Testing} & \multirow{2}{*}{\begin{tabular}[c]{@{}c@{}}\ \ \ ViLT\ \ \ \\ \cite{kim2021vilt}\end{tabular} } & \multirow{2}{*}{\begin{tabular}[c]{@{}c@{}}Lee et al. \\ \cite{Lee_2023_CVPR}\end{tabular}} & \multirow{2}{*}{\begin{tabular}[c]{@{}c@{}}Unimodal \\ Baseline\end{tabular}} & \multirow{2}{*}{\ \ \  Ours\ \ \ }  \\

                                                                                  & Image                  & Text                   & Image & Text                 &                            &                             &                                                                               & {}                       \\ 
\hline\hline

\multirow{12}{*}{\begin{tabular}[c]{@{}c@{}}MM-IMDB\\ (F1-Macro)\end{tabular}}    & \multirow{4}{*}{100\%} & \multirow{4}{*}{30\%}  & 100\% & 30\%                 & 32.78                      & 37.72                       & 39.47                                                                         & \textbf{43.21}                              \\
                                                                                  &                        &                        & 30\%  & 100\%                & 26.58                      & 21.68                       & 42.12                                                                         & \textbf{54.67}                              \\
                                                                                  &                        &                        & 65\%  & 65\%                 & 30.55                      & 30.80                       & 40.99                                                                         & \textbf{49.09}                              \\ \cline{4-9}
                                                                                  &                        &                        & \multicolumn{2}{c|}{ AVG}     & 29.97 &  30.07&  40.86&  {\textbf{48.99}}          \\ 
\cline{2-9}
                                                                                  & \multirow{4}{*}{30\%}  & \multirow{4}{*}{100\%} & 100\% & 30\%                 & 30.25                      & 24.93                       & 29.85                                                                         & \textbf{43.07}                              \\
                                                                                  &                        &                        & 30\%  & 100\%                & 37.97                      & 47.10                       & 54.37                                                                         & \textbf{56.03}                              \\
                                                                                  &                        &                        & 65\%  & 65\%                 & 34.45                      & 36.76                       & 37.61                                                                         & \textbf{49.81}                              \\\cline{4-9}
                                                                                  &                        &                        & \multicolumn{2}{c|}{ AVG}     &  34.22&  36.26&  40.61&  {\textbf{49.64}}          \\ 
\cline{2-9}
                                                                                  & \multirow{4}{*}{65\%}  & \multirow{4}{*}{65\%}  & 100\% & 30\%                 & 35.80                      & 39.04                       & 40.60                                                                         & \textbf{42.46}                              \\
                                                                                  &                        &                        & 30\%  & 100\%                & 36.65                      & 42.68                       & 53.19                                                                         & \textbf{55.26}                              \\
                                                                                  &                        &                        & 65\%  & 65\%                 & 36.66                      & 41.33                       & 47.34                                                                         & \textbf{49.24}                              \\\cline{4-9}
                                                                                  &                        &                        & \multicolumn{2}{c|}{ AVG}     &  {36.37} &  41.02&  47.04&  {\textbf{48.99}}          \\ 
\hline
\multirow{12}{*}{\begin{tabular}[c]{@{}c@{}}Food-101\\ (Accuracy)\end{tabular}}   & \multirow{4}{*}{100\%} & \multirow{4}{*}{30\%}  & 100\% & 30\%                 & 66.02                      & 73.51                       & 78.60                                                                         & \textbf{78.81}                              \\
                                                                                  &                        &                        & 30\%  & 100\%                & 42.89                      & 27.55                       & 82.50                                                                         & \textbf{86.90}                              \\
                                                                                  &                        &                        & 65\%  & 65\%                 & 54.49                      & 50.59                       & 80.37                                                                         & \textbf{82.35}                              \\\cline{4-9}
                                                                                  &                        &                        & \multicolumn{2}{c|}{ AVG}     &  {54.47} &  {50.55}  &  {80.49}                                                    &  {\textbf{82.69}}          \\ 
\cline{2-9}
                                                                                  & \multirow{4}{*}{30\%}  & \multirow{4}{*}{100\%} & 100\% & 30\%                 & 42.80                      & 29.61                       & 66.50                                                                         & \textbf{75.41}                              \\
                                                                                  &                        &                        & 30\%  & 100\%                & 76.54                      & 86.45                       & \textbf{87.44}                                                                & 87.11                                       \\
                                                                                  &                        &                        & 65\%  & 65\%                 & 59.54                      & 57.97                       & 76.98                                                                         & \textbf{81.22}                              \\\cline{4-9}
                                                                                  &                        &                        & \multicolumn{2}{c|}{ AVG}     &  {59.63} &  {58.01}  &  {76.97}                                                    &  {\textbf{81.25}}          \\ 
\cline{2-9}
                                                                                  & \multirow{4}{*}{65\%}  & \multirow{4}{*}{65\%}  & 100\% & 30\%                 & 64.40                      & 71.26                       & 76.51                                                                         & \textbf{78.13}                              \\
                                                                                  &                        &                        & 30\%  & 100\%                & 73.60                      & 85.71                       & 87.21                                                                         & \textbf{87.35}                              \\
                                                                                  &                        &                        & 65\%  & 65\%                 & 68.96                      & 78.43                       & 81.82                                                                         & \textbf{82.67}                              \\\cline{4-9}
                                                                                  &                        &                        & \multicolumn{2}{c|}{ AVG}     &  {68.99} &  {78.47}  &  {81.85}                                                    &  {\textbf{82.72}}          \\ 
\hline
\multirow{12}{*}{\begin{tabular}[c]{@{}c@{}}Hateful Memes\\ (AUROC)\end{tabular}} & \multirow{4}{*}{100\%} & \multirow{4}{*}{30\%}  & 100\% & 30\%                 & 60.77                      & 58.54                       & 59.77                                                                         & \textbf{61.39}                              \\
                                                                                  &                        &                        & 30\%  & 100\%                & 61.84                      & 55.67                       & 66.41                                                                         & \textbf{66.50}                              \\
                                                                                  &                        &                        & 65\%  & 65\%                 & 61.57                      & 57.10                       & 61.86                                                                         & \textbf{63.35}                              \\\cline{4-9}
                                                                                  &                        &                        & \multicolumn{2}{c|}{ AVG}     &  {61.39} &  {57.10}  &  {62.68}                                                    &  {\textbf{63.75}}          \\ 
\cline{2-9}
                                                                                  & \multirow{4}{*}{30\%}  & \multirow{4}{*}{100\%} & 100\% & 30\%                 & 56.19                      & 58.43                       & 58.98& \textbf{62.06}                              \\
                                                                                  &                        &                        & 30\%  & 100\%                & 62.78                      & 65.54                       & 68.52                                                                         & \textbf{68.97}                              \\
                                                                                  &                        &                        & 65\%  & 65\%                 & 59.29                      & 62.36                       & 63.58                                                                         & \textbf{65.12}                              \\\cline{4-9}
                                                                                  &                        &                        & \multicolumn{2}{c|}{ AVG}     &  {59.42} &  {62.11}  &  {63.69}&  {\textbf{65.38}}          \\ 
\cline{2-9}
                                                                                  & \multirow{4}{*}{65\%}  & \multirow{4}{*}{65\%}  & 100\% & 30\%                 & 59.42                      & 60.05                       & 60.93                                                                         & \textbf{61.39}                              \\
                                                                                  &                        &                        & 30\%  & 100\%                & 63.13                      & 61.88                       & \textbf{69.19}                                                                & 68.24                                       \\
                                                                                  &                        &                        & 65\%  & 65\%                 & 61.32                      & 61.52                       & 64.47                                                                         & \textbf{65.17}                              \\\cline{4-9}
                                                                                  &                        &                        & \multicolumn{2}{c|}{ AVG}     &  {61.29} &  {61.15}  &  {64.86}                                                    &  {\textbf{64.93}}         \\
 \bottomrule      
\end{tabular}}
\end{table*}
Fig. \ref{fig:result1} shows performance under the complete training setting, highlighting declines when text is missing at inference. Although all methods perform similarly well when all modalities are present, they struggle when a dominant modality is absent during testing. This result aligns with a prior study \cite{Ma_2022_CVPR} that the performance of multimodal models trained on complete data degrades when faced with incomplete data at inference. Our findings indicate that using separate unimodal encoders for multimodal learning is also susceptible to missing modalities. Specifically, on MM-IMDb and Food-101, the performance of the unimodal baseline is even worse than that of training solely with the image encoder when less than 10\% of the test data are paired. Our approach, however, stands out by significantly outperforming others, especially when the text is severely missing.

Moreover, on the Hateful Memes, the performance gap between the unimodal baseline and the image-encoder-only approach is slight, indicating that it does not rely on a single modality. As shown, our model is always superior to others.

\begin{minipage}[t!]{0.4\textwidth}\
    \includegraphics[width=\linewidth]{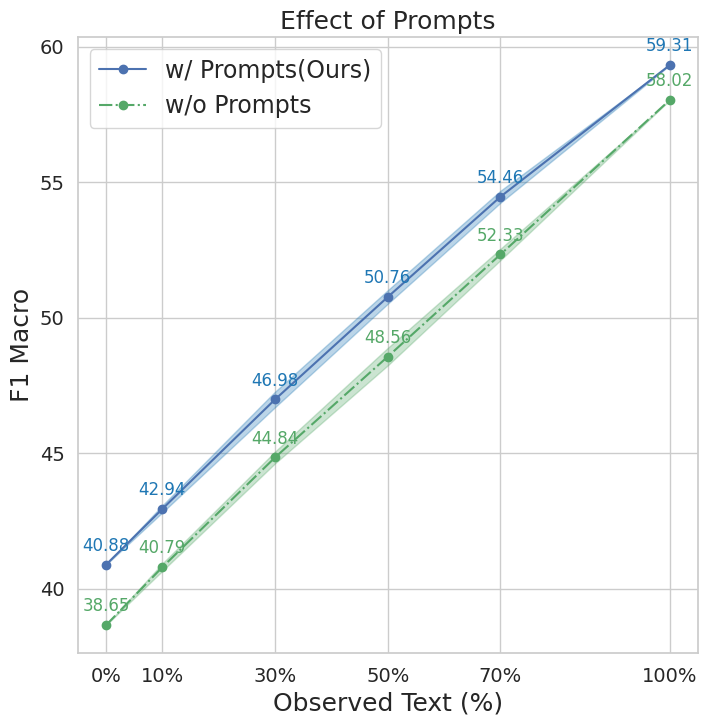}\
    \captionof{figure}{Ablation on effect of the prompts under complete training setting.}
    \label{fig:prompts_effect_on_complete}
\end{minipage}
\hfill
\begin{minipage}[t!]{0.45\textwidth}
\captionof{table}{Ablation study on effect of the prompts under missing training setting. The results are averaged over three distinct missing cases, each conducted with five different random seeds. }
\begin{tabular}{cc|m{1.5cm}|m{1.5cm}}
 \toprule
\multicolumn{2}{c|}{Training} & \multirow{2}{*}{\begin{tabular}[c]{@{}c@{}}w/o\\ \ Prompts\ \end{tabular}} & \multirow{2}{*}
{\begin{tabular}[c]{@{}c@{}}\ \ \ Ours\ \end{tabular}}  \\
Image         & Text          &                                                                        &                       \\ \hline \hline
100\%         & 30\%          & \multicolumn{1}{c|}{44.91}                                                                  & \multicolumn{1}{c}{\textbf{48.99}}        \\
30\%          & 100\%         & \multicolumn{1}{c|}{45.46}                                                                  & \multicolumn{1}{c}{\textbf{49.64}}        \\
65\%          & 65\%          & \multicolumn{1}{c|}{46.98}                                                                  & \multicolumn{1}{c}{\textbf{48.99}}       \\ \bottomrule
\end{tabular}
\label{tab:prompts_effect_on_missing}
\end{minipage}

\noindent It achieves an AUROC of 60.4, which is 6.5\% greater than unimodal baseline AUROC of 56.73, providing that our predictor modules with read-only prompts generate auxiliary text representations that prove beneficial for the target task at hand. Overall, our method outperforms state-of-the-art performance across various datasets and evaluation metrics in the severely missing cases, and it is even better than multimodal models pretrained on large image-text pairs.

The results for the second setting, as shown in Table \ref{missing_result}, display performance across different missing scenarios with 70\% of modalities missing. Each training configuration is evaluated (\ie, testing configuration) in a scenario where the modalities are missing equally, as in the training phase, and in two additional cases that differ from the training scenario. As shown, we observe a lack of robustness in other methods when faced with missing scenarios that significantly deviate from the training settings. For instance, models trained with 30\% image and 100\% text samples demonstrate adequate performance in test samples with an equal distribution of missing modalities. However, their performance significantly degrades when encountering samples with 100\% image and 30\% text. Prior work \cite{Lee_2023_CVPR} using multimodal encoder especially does not cope with unseen cases because it did not experience the missing setting during training. They even demonstrate inferior performance compared to ViLT. While unimodal baseline outperforms our method in a few specific scenarios where the missing modalities in training and testing are aligned, it is limited in such conditions and susceptible to different missing scenarios.

Conversely, our method maintains robustness to missing modalities, even when substantial differences exist in the train-test settings. As a result, when averaging the performances of three different missing cases, our method outperforms the state-of-the-art by a large margin in various datasets and settings because of leveraging pretrained knowledge of unimodal models and predictor modules. Importantly, it achieves competitive results despite the limited (30\%) availability of paired data and the absence of multimodal pretraining (\ie, a multimodal joint encoder), underscoring its robustness, applicability, and flexibility in real-world scenarios.
\subsection{Ablation Study}
\subsubsection{Effect of Prompts-based Feature Prediction}\label{sec:ablation_effect_prompts} 
\begin{figure}[tp]
  \centering
  \begin{subfigure}{0.48\linewidth}
    \centering
    \includegraphics[width=\linewidth]{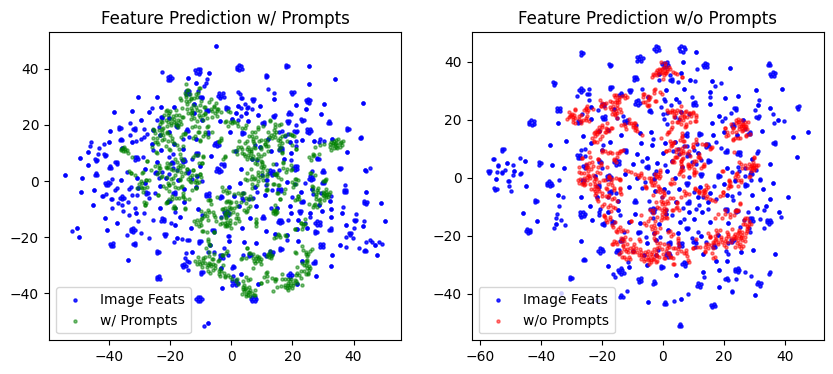}  
    \caption{Image feature prediction}
    \label{fig:half-a}
  \end{subfigure}
  \hfill
  \begin{subfigure}{0.48\linewidth}
    \centering
    \includegraphics[width=\linewidth]{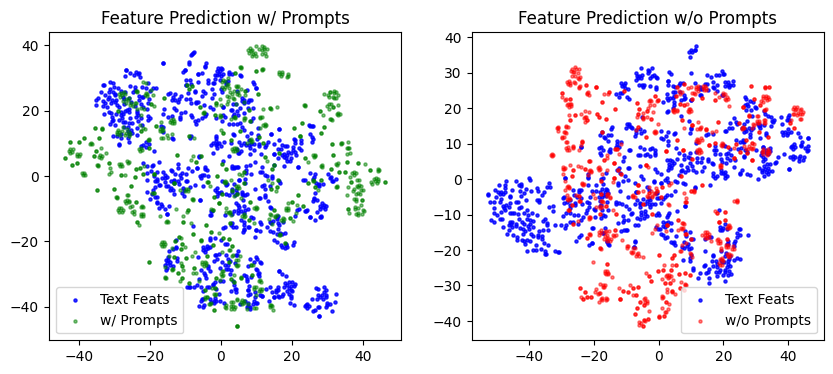}   
    \caption{Text feature prediction}
    \label{fig:half-b}
  \end{subfigure}
  \caption{ \textbf{t-SNE visualization of (a) image feature, (b) text feature.} Feature prediction with prompts (green) results in embeddings that align more closely with the ground truth features (blue), compared to prediction without prompts (red).}
  \label{fig:tsne}
\end{figure}

We investigate the effects of prompts for feature prediction on the MM-IMDb, as shown in Fig. \ref{fig:prompts_effect_on_complete} and Table \ref{tab:prompts_effect_on_missing}. We compare our method with a predictor leveraging the CLS token, which captures the aggregated information of input sequences. Our prompts-based method only requires an additional 0.005\% parameters of backbone but outperforms the method leveraging CLS token for feature prediction in both scenarios. It reveals that the CLS token is specialized for the target task and less suitable for predicting the embedding of other modalities. In the next step, we present the t-SNE \cite{tsne} visualizations of the embeddings in Fig. \ref{fig:tsne}. Each data point in the figure represents ground truth embeddings, and the prediction leveraging read-only prompts and CLS token for each test sample. The figure illustrates that our method produces embeddings more closely aligned with the original features than using the CLS token. Finally, we examine the cosine similarity of our feature predictions for quantitative confirmation. Surprisingly, the prompts-based approach increased the similarity of text prediction from 0.54 to 0.57 and improved image feature prediction from 0.4 to 0.71. 

\begin{figure*}[tp]
    \centering
    \includegraphics[width=0.5\textwidth]{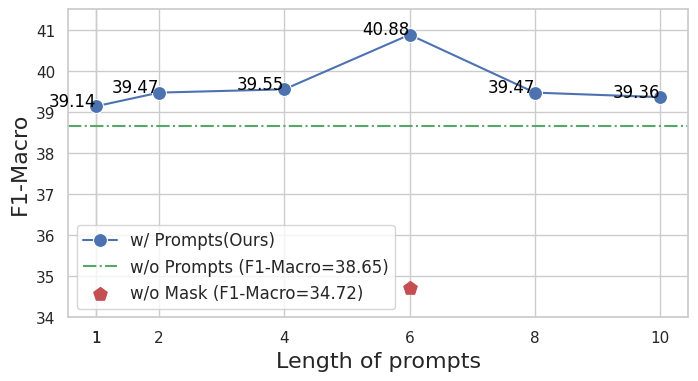}
    \caption{Ablation study on different length of prompts and attention masking. }
    \label{fig:prompt_length}
\end{figure*}

\begin{minipage}[tp]{0.4\textwidth}
    \includegraphics[width=\linewidth]{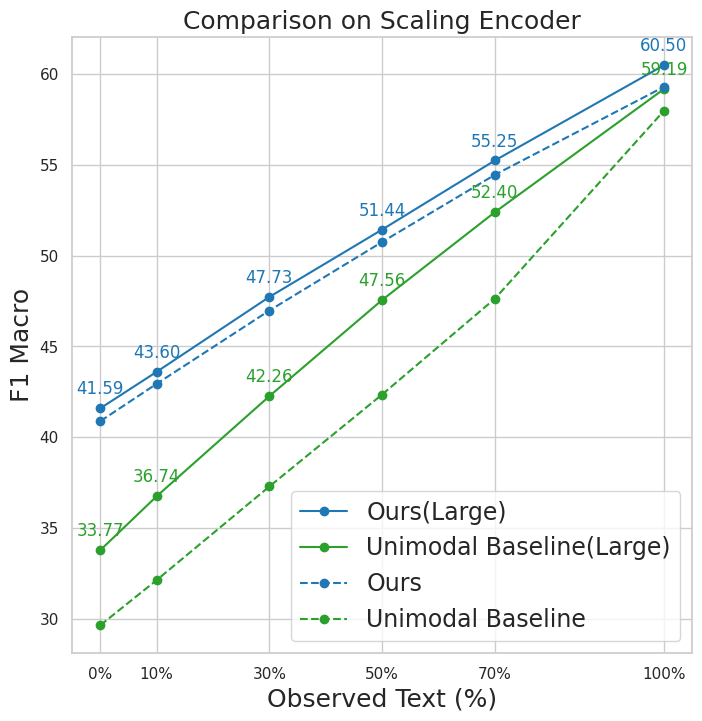}
    \captionof{figure}{Ablation on scaling encoders under complete training.}
    \label{fig:scaling}
\end{minipage}
\hfill
\begin{minipage}[tp]{0.45\textwidth}
\captionof{table}{ Impact of variance-covariance regularization. Inv: a invariance loss is used, Var: variance regularization, Cov: covariance regularization, in Eq \ref{eq:vicreg}. }
\begin{tabular}{l|cl}
\toprule 
\multicolumn{1}{l|}{Method} & \multicolumn{2}{c}{F1-Macro} \\ \hline \hline
Inv w/ stop gradient                                      & \multicolumn{2}{c}{22.72}         \\ \hline
Inv                                      & \multicolumn{2}{c}{25.72}         \\ 
\multicolumn{1}{l|}{Inv+Var+Cov (VICReg)} & \multicolumn{2}{c}{40.88}         \\ \bottomrule
\end{tabular}
\label{tab:vicreg}
\end{minipage}
\subsubsection{Further Analysis of Read-only Prompts} 
 We conduct an ablation experiment to analyze the effect of prompt token length under the complete training setting, as illustrated in Fig. \ref{fig:prompt_length}. The performance evaluated with text-missing data improves with an increase in prompt length, peaking at a length of 6, beyond which additional length does not proportionally enhance performance. Interestingly, compared to using the CLS token, we observe meaningful improvement when leveraging even one learnable prompt token, demonstrating the effectiveness of prompt-based feature prediction. This aligns with findings from the previous section that using additional prompt tokens instead of CLS token, specialized for downstream tasks, enhances feature prediction.

Furthermore, to assess the effect of the read-only mechanism, we examined its performance using prompts without attention masking. The figure shows a dramatic decrease in performance to 34.72 from 40.88 in an equal prompt setting. It emphasizes the necessity of structuring read-only prompts in predicting features to preserve the model's internal representations for the target task.
\subsubsection{Scaling} 
As illustrated in Fig. \ref{fig:scaling}, we explicitly explore the performance enhancements achieved by employing advanced unimodal encoders, particularly a scaled-up version of our base encoder, termed the "Large". The aim here is twofold: first, our method outperforms unimodal baselines even with the larger encoder, and second, it underscores the advantages of using unimodal pretrained encoders in real-world scenarios. Our results indicate that more powerful encoders enhance performance and improve robustness in scenarios with missing modalities. Unimodal pretrained encoders, often more accessible and available in high-performance variants, present a more feasible option for enhancing model capabilities than multimodal encoders. This highlights our method's adaptability and efficiency in real-world scenarios, especially with domain-specific datasets. By incorporating domain-specific encoders, such as multilingual pretrained models, we expect even more significant performance improvements, showcasing our approach's potential to effectively leverage advancements in unimodal encoding.

\subsubsection{Impact of VICReg}
Table \ref{tab:vicreg} presents the results of text-missing testing under complete training, highlighting VICReg's crucial role in our method. It indicates that the stop-gradient operation is not required, which restricts the predictability of encoders. Notably, relying solely on the invariance term, the primary objective for feature prediction, resulted in a marked decline. Encoders are trained to make embeddings easily predictable from other modalities by simply adding the covariance and variance regularization terms. These results demonstrate how VICReg significantly enhances predictability, guiding the model toward producing informative representations.

\subsubsection{Comparison on other PEFT methods}\label{abl:peft}
To assess the compatibility of our framework with other PEFT methods, we conduct a comparative analysis with Layer Normalization (LN) Tuning \cite{lntuning}, Prefix Tuning \cite{li2021prefix}, and Adapter Tuning \cite{houlsby2019parameter} under complete training setting. 
Table \ref{tab:abl_peft} presents the performance comparison on text-missing data. Our observations reveal that BitFit surpasses other methods with very few (0.11\%) trainable parameters. Despite having the most trainable parameters, adapter-based methods underperformed, indicating that slightly tuning the encoders across all layers is enough to train the encoder's predictability while optimizing the target task and often offers more benefits than focusing on specific layers.
\begin{table}[t]
\caption{Ablation study on other PEFT methods. "\% Trained" indicates the ratio of trainable parameters to the total parameters in fully fine-tuned encoders. † The "Adapter" was inserted in three configurations: first layer, last layer, and layerwise.}
\centering
\begin{tabular*}{\textwidth}{@{\extracolsep{\fill}} lcc}
\toprule %
Method         & \% Trained & F1-Macro \\ \hline  \hline
BitFit(Ours) \cite{zaken2021bitfit}   &   0.11         & \textbf{40.88}    \\
LN Tuning \cite{lntuning}      &   0.04         & 39.07    \\
Prefix Tuning \cite{li2021prefix} &   0.03         & 37.75    \\ 
Adapter † (first/last/layerwise) \cite{houlsby2019parameter} &   0.60/0.60/6.80         &    30.83/34.42/37.37\\ 
\bottomrule %
\end{tabular*}
\label{tab:abl_peft}
\end{table}

\section{Conclusion}
This paper addresses the practical challenges in multimodal learning associated with acquiring complete multimodal data. In real-world scenarios, using a pretrained joint encoder on a large paired dataset may not always be feasible. Furthermore, the issue missing modalities for downstream tasks presents potential challenges during both training~(fine-tuning) and testing phases. We introduce a simple yet effective framework designed to tackle missing modalities by employing PEFT on separate pretrained unimodal models. Our approach utilizes VICReg to effectively predict the embeddings of other modality within the representation space, leveraging read-only prompts. Our method exhibits superior performance across different multimodal datasets in various scenarios for missing modality that occurs during both the training and testing phases.

\section*{Acknowledgements} 
This work was supported by the National Research Foundation of Korea (NRF) grant (RS-2023-00222663, RS-2024-00345809) and Institute of Information \& communications Technology Planning \& Evaluation (IITP) grant (RS-2022-00143911, RS2023-00232046, IITP-2024-RS-2024-00397085), both funded by the Korea government (MSIT). 
%
%
\bibliography{main}

\begin{thebibliography}{10}
\providecommand{\url}[1]{\texttt{#1}}
\providecommand{\urlprefix}{URL }
\providecommand{\doi}[1]{https://doi.org/#1}

\bibitem{arevalo2017gated}
Arevalo, J., Solorio, T., Montes-y G{\'o}mez, M., Gonz{\'a}lez, F.A.: Gated multimodal units for information fusion. arXiv preprint arXiv:1702.01992  (2017)

\bibitem{ba2016layer}
Ba, J.L., Kiros, J.R., Hinton, G.E.: Layer normalization. arXiv preprint arXiv:1607.06450  (2016)

\bibitem{bardes2021vicreg}
Bardes, A., Ponce, J., LeCun, Y.: Vicreg: Variance-invariance-covariance regularization for self-supervised learning. arXiv preprint arXiv:2105.04906  (2021)

\bibitem{swav}
Caron, M., Misra, I., Mairal, J., Goyal, P., Bojanowski, P., Joulin, A.: Unsupervised learning of visual features by contrasting cluster assignments. Advances in neural information processing systems  \textbf{33},  9912--9924 (2020)

\bibitem{SimCLR}
Chen, T., Kornblith, S., Norouzi, M., Hinton, G.: A simple framework for contrastive learning of visual representations. In: International conference on machine learning. pp. 1597--1607. PMLR (2020)

\bibitem{SimSiam}
Chen, X., He, K.: Exploring simple siamese representation learning. In: Proceedings of the IEEE/CVF conference on computer vision and pattern recognition. pp. 15750--15758 (2021)

\bibitem{chen2020uniter}
Chen, Y.C., Li, L., Yu, L., El~Kholy, A., Ahmed, F., Gan, Z., Cheng, Y., Liu, J.: Uniter: Universal image-text representation learning. In: European conference on computer vision. pp. 104--120. Springer (2020)

\bibitem{vit}
Dosovitskiy, A., Beyer, L., Kolesnikov, A., Weissenborn, D., Zhai, X., Unterthiner, T., Dehghani, M., Minderer, M., Heigold, G., Gelly, S., et~al.: An image is worth 16x16 words: Transformers for image recognition at scale. arXiv preprint arXiv:2010.11929  (2020)

\bibitem{gao2021simcse}
Gao, T., Yao, X., Chen, D.: Simcse: Simple contrastive learning of sentence embeddings. arXiv preprint arXiv:2104.08821  (2021)

\bibitem{audio-visual}
Gong, Y., Rouditchenko, A., Liu, A.H., Harwath, D., Karlinsky, L., Kuehne, H., Glass, J.: Contrastive audio-visual masked autoencoder. arXiv preprint arXiv:2210.07839  (2022)

\bibitem{BYOL}
Grill, J.B., Strub, F., Altch{\'e}, F., Tallec, C., Richemond, P., Buchatskaya, E., Doersch, C., Avila~Pires, B., Guo, Z., Gheshlaghi~Azar, M., et~al.: Bootstrap your own latent-a new approach to self-supervised learning. Advances in neural information processing systems  \textbf{33},  21271--21284 (2020)

\bibitem{houlsby2019parameter}
Houlsby, N., Giurgiu, A., Jastrzebski, S., Morrone, B., De~Laroussilhe, Q., Gesmundo, A., Attariyan, M., Gelly, S.: Parameter-efficient transfer learning for nlp. In: International Conference on Machine Learning. pp. 2790--2799. PMLR (2019)

\bibitem{hu2021lora}
Hu, E.J., Shen, Y., Wallis, P., Allen-Zhu, Z., Li, Y., Wang, S., Wang, L., Chen, W.: Lora: Low-rank adaptation of large language models. arXiv preprint arXiv:2106.09685  (2021)

\bibitem{medical2}
Huang, S.C., Shen, L., Lungren, M.P., Yeung, S.: Gloria: A multimodal global-local representation learning framework for label-efficient medical image recognition. In: Proceedings of the IEEE/CVF International Conference on Computer Vision. pp. 3942--3951 (2021)

\bibitem{vpt}
Jia, M., Tang, L., Chen, B.C., Cardie, C., Belongie, S., Hariharan, B., Lim, S.N.: Visual prompt tuning. In: European Conference on Computer Vision. pp. 709--727. Springer (2022)

\bibitem{khan2023contrastive}
Khan, Z., Fu, Y.: Contrastive alignment of vision to language through parameter-efficient transfer learning. arXiv preprint arXiv:2303.11866  (2023)

\bibitem{khattak2023maple}
Khattak, M.U., Rasheed, H., Maaz, M., Khan, S., Khan, F.S.: Maple: Multi-modal prompt learning. In: Proceedings of the IEEE/CVF Conference on Computer Vision and Pattern Recognition. pp. 19113--19122 (2023)

\bibitem{promptsrc}
Khattak, M.U., Wasim, S.T., Naseer, M., Khan, S., Yang, M.H., Khan, F.S.: Self-regulating prompts: Foundational model adaptation without forgetting. In: Proceedings of the IEEE/CVF International Conference on Computer Vision. pp. 15190--15200 (2023)

\bibitem{kiela2020hateful}
Kiela, D., Firooz, H., Mohan, A., Goswami, V., Singh, A., Ringshia, P., Testuggine, D.: The hateful memes challenge: Detecting hate speech in multimodal memes. Advances in neural information processing systems  \textbf{33},  2611--2624 (2020)

\bibitem{kim2021vilt}
Kim, W., Son, B., Kim, I.: Vilt: Vision-and-language transformer without convolution or region supervision. In: International Conference on Machine Learning. pp. 5583--5594. PMLR (2021)

\bibitem{Multilingualfashion}
Kosar, V., Hoskovec, A., {\v{S}}ulc, M., Bartyzal, R.: Glami-1m: A multilingual image-text fashion dataset. arXiv preprint arXiv:2211.14451  (2022)

\bibitem{lecun2022path}
LeCun, Y.: A path towards autonomous machine intelligence version 0.9. 2, 2022-06-27. Open Review  \textbf{62}(1) (2022)

\bibitem{lee2023read}
Lee, D., Song, S., Suh, J., Choi, J., Lee, S., Kim, H.J.: Read-only prompt optimization for vision-language few-shot learning. In: Proceedings of the IEEE/CVF International Conference on Computer Vision. pp. 1401--1411 (2023)

\bibitem{Lee_2023_CVPR}
Lee, Y.L., Tsai, Y.H., Chiu, W.C., Lee, C.Y.: Multimodal prompting with missing modalities for visual recognition. In: Proceedings of the IEEE/CVF Conference on Computer Vision and Pattern Recognition (CVPR). pp. 14943--14952 (June 2023)

\bibitem{prompt}
Lester, B., Al-Rfou, R., Constant, N.: The power of scale for parameter-efficient prompt tuning. arXiv preprint arXiv:2104.08691  (2021)

\bibitem{li2021prefix}
Li, X.L., Liang, P.: Prefix-tuning: Optimizing continuous prompts for generation. arXiv preprint arXiv:2101.00190  (2021)

\bibitem{loshchilov2018fixing}
Loshchilov, I., Hutter, F.: Fixing weight decay regularization in adam  (2018)

\bibitem{Ma_2022_CVPR}
Ma, M., Ren, J., Zhao, L., Testuggine, D., Peng, X.: Are multimodal transformers robust to missing modality? In: Proceedings of the IEEE/CVF Conference on Computer Vision and Pattern Recognition (CVPR). pp. 18177--18186 (June 2022)

\bibitem{ma2021smil}
Ma, M., Ren, J., Zhao, L., Tulyakov, S., Wu, C., Peng, X.: Smil: Multimodal learning with severely missing modality. In: Proceedings of the AAAI Conference on Artificial Intelligence. vol.~35, pp. 2302--2310 (2021)

\bibitem{tsne}
Van~der Maaten, L., Hinton, G.: Visualizing data using t-sne. Journal of machine learning research  \textbf{9}(11) (2008)

\bibitem{lntuning}
Qi, W., Ruan, Y.P., Zuo, Y., Li, T.: Parameter-efficient tuning on layer normalization for pre-trained language models. arXiv preprint arXiv:2211.08682  (2022)

\bibitem{clip}
Radford, A., Kim, J.W., Hallacy, C., Ramesh, A., Goh, G., Agarwal, S., Sastry, G., Askell, A., Mishkin, P., Clark, J., et~al.: Learning transferable visual models from natural language supervision. In: International conference on machine learning. pp. 8748--8763. PMLR (2021)

\bibitem{shwartz2023information}
Shwartz-Ziv, R., Balestriero, R., Kawaguchi, K., Rudner, T.G., LeCun, Y.: An information-theoretic perspective on variance-invariance-covariance regularization. arXiv preprint arXiv:2303.00633  (2023)

\bibitem{multilingaulwiki}
Srinivasan, K., Raman, K., Chen, J., Bendersky, M., Najork, M.: Wit: Wikipedia-based image text dataset for multimodal multilingual machine learning. In: Proceedings of the 44th International ACM SIGIR Conference on Research and Development in Information Retrieval. pp. 2443--2449 (2021)

\bibitem{sun2019videobert}
Sun, C., Myers, A., Vondrick, C., Murphy, K., Schmid, C.: Videobert: A joint model for video and language representation learning. In: Proceedings of the IEEE/CVF international conference on computer vision. pp. 7464--7473 (2019)

\bibitem{tan2019lxmert}
Tan, H., Bansal, M.: Lxmert: Learning cross-modality encoder representations from transformers. arXiv preprint arXiv:1908.07490  (2019)

\bibitem{touvron2022deit}
Touvron, H., Cord, M., J{\'e}gou, H.: Deit iii: Revenge of the vit. In: European Conference on Computer Vision. pp. 516--533. Springer (2022)

\bibitem{vaswani2017attention}
Vaswani, A., Shazeer, N., Parmar, N., Uszkoreit, J., Jones, L., Gomez, A.N., Kaiser, {\L}., Polosukhin, I.: Attention is all you need. Advances in neural information processing systems  \textbf{30} (2017)

\bibitem{wang2023multi}
Wang, H., Chen, Y., Ma, C., Avery, J., Hull, L., Carneiro, G.: Multi-modal learning with missing modality via shared-specific feature modelling. In: Proceedings of the IEEE/CVF Conference on Computer Vision and Pattern Recognition. pp. 15878--15887 (2023)

\bibitem{wang2015recipe}
Wang, X., Kumar, D., Thome, N., Cord, M., Precioso, F.: Recipe recognition with large multimodal food dataset. In: 2015 IEEE International Conference on Multimedia \& Expo Workshops (ICMEW). pp.~1--6. IEEE (2015)

\bibitem{sprompt}
Wang, Y., Huang, Z., Hong, X.: S-prompts learning with pre-trained transformers: An occam’s razor for domain incremental learning. Advances in Neural Information Processing Systems  \textbf{35},  5682--5695 (2022)

\bibitem{wang2022dualprompt}
Wang, Z., Zhang, Z., Ebrahimi, S., Sun, R., Zhang, H., Lee, C.Y., Ren, X., Su, G., Perot, V., Dy, J., et~al.: Dualprompt: Complementary prompting for rehearsal-free continual learning. In: European Conference on Computer Vision. pp. 631--648. Springer (2022)

\bibitem{l2p}
Wang, Z., Zhang, Z., Lee, C.Y., Zhang, H., Sun, R., Ren, X., Su, G., Perot, V., Dy, J., Pfister, T.: Learning to prompt for continual learning. In: Proceedings of the IEEE/CVF Conference on Computer Vision and Pattern Recognition. pp. 139--149 (2022)

\bibitem{zaken2021bitfit}
Zaken, E.B., Ravfogel, S., Goldberg, Y.: Bitfit: Simple parameter-efficient fine-tuning for transformer-based masked language-models. arXiv preprint arXiv:2106.10199  (2021)

\bibitem{medical1}
Zhang, Y., Jiang, H., Miura, Y., Manning, C.D., Langlotz, C.P.: Contrastive learning of medical visual representations from paired images and text. In: Machine Learning for Healthcare Conference. pp. 2--25. PMLR (2022)

\bibitem{cocoop}
Zhou, K., Yang, J., Loy, C.C., Liu, Z.: Conditional prompt learning for vision-language models. In: Proceedings of the IEEE/CVF Conference on Computer Vision and Pattern Recognition. pp. 16816--16825 (2022)

\bibitem{coop}
Zhou, K., Yang, J., Loy, C.C., Liu, Z.: Learning to prompt for vision-language models. International Journal of Computer Vision  \textbf{130}(9),  2337--2348 (2022)

\end{thebibliography}

\clearpage
\appendix
\section*{\centering Supplementary: Missing Modality Prediction \\ for Unpaired Multimodal Learning \\ via Joint Embedding of Unimodal Models}

\renewcommand{\thesection}{\Alph{section}}
\setcounter{section}{0}

\section{Implementation Details}
We conduct overall experiments using a single RTX 3090 GPU. Similar to \cite{khan2023contrastive}, we employ DeiT III \cite{touvron2022deit} as an image encoder and SimCSE \cite{gao2021simcse} as a text encoder throughout the overall experiment, which model parameters are initialized using the pretrained weights. The structure of the feature predictor is modified from \cite{bardes2021vicreg}, which consists of two fully-connected layers with layer normalization layer \cite{ba2016layer} and activation function, and a third linear layer. The dimensions of all three layers are set to equal to the output dimension of the encoders. The length of learnable read-only prompts for feature prediction is set to 6 in MM-IMDb, 20 in UPMC Food-101, and 2 in Hateful Memes in both training settings. We conduct all experiments with batch size 12 of 20 epochs, using the AdamW \cite{loshchilov2018fixing} optimizer with a weight decay of $5 \times 10^{-2}$. We initiate the warm-up steps for the learning rate at 0, with a base learning rate set to $1 \times 10^{-2}$. The warm-up phase linearly progresses from 0 for the first 10\% of the total training steps before decay. The variance, invariance, and covariance coefficients $\lambda, \mu$ and $\nu$ for VICReg loss in Eq \ref{eq:vicreg} are set to 50, 50, and 1. 

For the ablation study on section \ref{abl:peft}, we add 36 learnable tokens to the input data specifically for the target task in prefix tuning, separate from utilizing read-only prompts for feature prediction. For adapter tuning, the adapter is inserted before each layer normalization layer in transformer-based pretrained models. It consists of a weight matrix that downsamples the input, followed by an activation function, and another weight matrix that restores the input to its original dimension. Finally, it includes a residual connection. We set the reduction factor to 4. For a fair comparison that considers trainable parameters, adding an adapter is applied to the first and last layers, and separately to all layers for further analysis.
\clearpage

\section{Read Only Prompts Attention Mask}
\begin{figure*}[tp]
    \centering
    \includegraphics[width=0.5\textwidth]{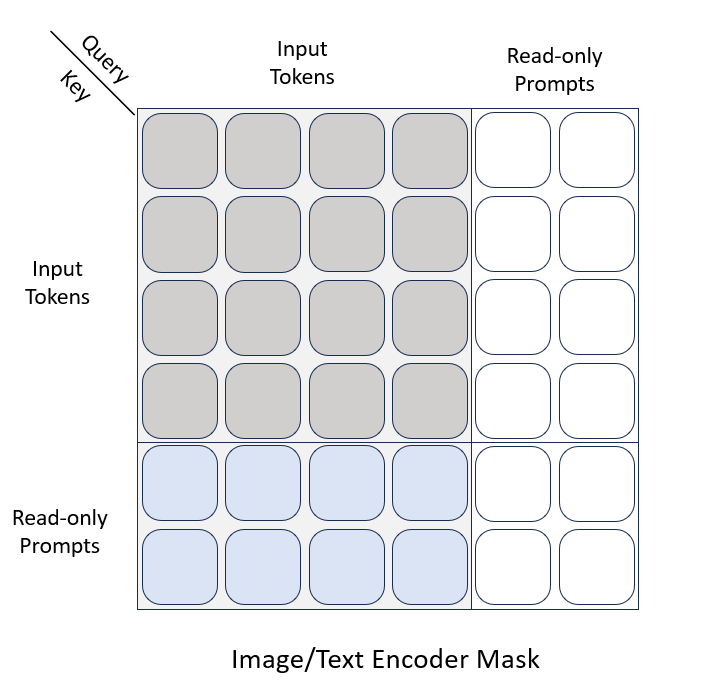}
    \caption{\textbf{The attention mask we use in Read only prompts}. The colored boxes indicate areas where attention operation occurs, while the white boxes indicate masked regions.}
    \label{fig:attention_mask}
\end{figure*}

As illustrated in Fig. \ref{fig:attention_mask}, we applied an attention masking mechanism for our encoder's architecture, following the approach outlined in \cite{lee2023read}. This structure is essential to maintain the representations of the input tokens for the target task while training the learnable tokens specialized for feature prediction.

\section{Additional Results}
\subsection{Results on Image Missing Case}
\begin{figure*}[tp]
    \centering
    \includegraphics[width=0.7\textwidth]{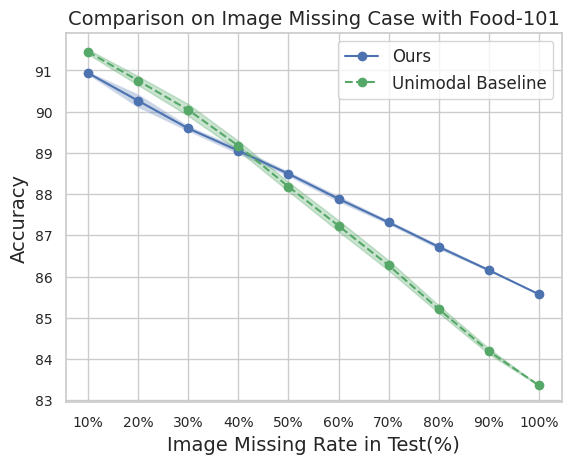}
    \caption{\textbf{Additional results on image missing case under complete training setting}. Dotted lines denote unimodal baseline, whereas solid lines represent ours.}
    \label{fig:image_missing}
\end{figure*}
We provide a result of an image missing case in the Fig \ref{fig:image_missing} under the complete training with the Food-101, which is the most susceptible among the three datasets. Due to the dominance of the text modality, the impact of the image modality on target task is minimal. However, the unimodal baseline suffers in severely image-missing cases. Conversely, our method, which demonstrates superior performance in the text missing case, also shows robustness to severely image missing cases

\subsection{Results on Different Missing Ratio}
\begin{figure*}[tp]
    \centering
    \includegraphics[width=0.7\textwidth]{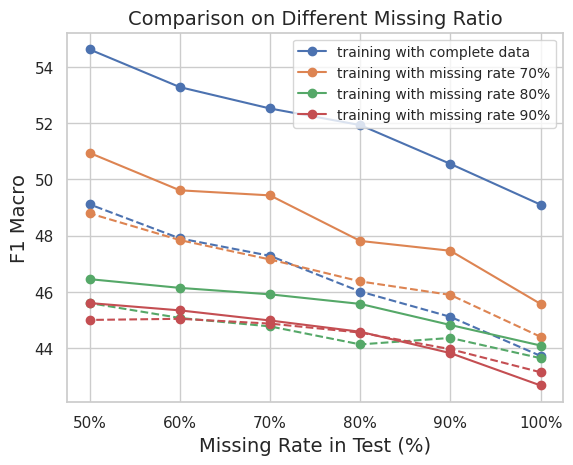}
    \caption{\textbf{Additional results on different missing ratio under missing training setting}. Dotted lines denote unimodal baseline, whereas solid lines represent ours}
    \label{fig:different_missing}
\end{figure*}

As illustrated in Fig. \ref{fig:different_missing}, we explored varying missing rates from the fixed 70\% on [both missing training/testing] setting. \textit{Our method with feature predictor trained on 70\% missing data (i.e., only 30\% paired) outperforms the unimodal baseline trained with fully-paired data.} When the missing rate is exceptionally high (i.e., 90\% missing), feature predictors struggle to learn effectively from the limited paired data available. Nonetheless, even with just 20\% paired data, our method surpasses the unimodal baseline.

\subsection{Full Results of the Ablation Study on prompts-based feature prediction}

\begin{figure*}[]
    \centering
    \includegraphics[width=\textwidth]{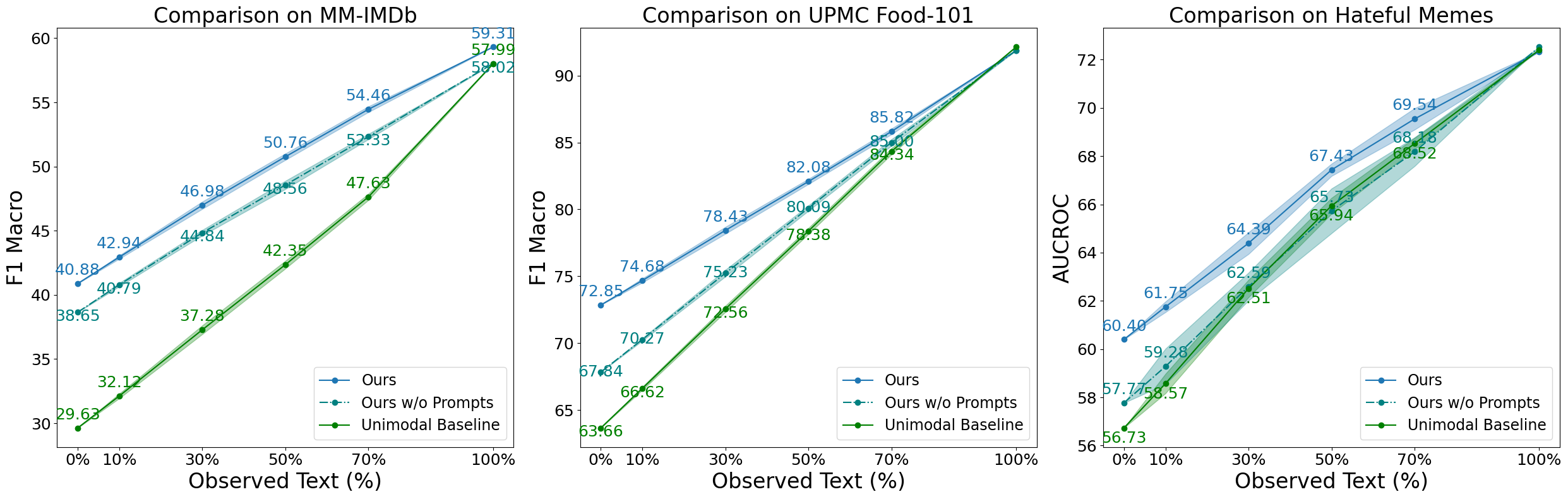}
    \caption{\textbf{Additional results on ablation study about the effect of prompts-based feature prediction under a complete training setting}. All experiments were conducted with 100\% text and 100\% image data and evaluated based on the text missing rate. For a fair comparison, we evaluate by averaging performances across five different random seeds. Ours w/o prompts leverage CLS token for feature prediction.}
    \label{fig:supple_figure}
\end{figure*}

\clearpage
\subsection{Full Results on Missing Training Setting}

\begin{table*}[]
\centering
\caption{
\textbf{Full comparison results under a training setting with a 70\% missing rate.} Evaluations were conducted by averaging the results from five different seeds across various modality-missing scenarios.}
\label{full_missing_result}
\resizebox{\textwidth}{!}{
\begin{tabular}{c|cc|cc|c|c|c|c|c} 
\toprule
\multirow{2}{*}{Dataset}                                                         & \multicolumn{2}{c|}{Training}                   & \multicolumn{2}{c|}{Testing} & \multirow{2}{*}{\ \ \ \ ViLT\ \ \ } & \multirow{2}{*}{Lee et al.} & \multirow{2}{*}{\begin{tabular}[c]{@{}c@{}}Unimodal \\ Baseline\end{tabular}} & \multirow{2}{*}{\begin{tabular}[c]{@{}c@{}}Ours w/o\\Prompts\end{tabular}}  & \multirow{2}{*}{\ \ \ \ Ours\ \ \ }  \\
                                                                                  & Image                  & Text                   & Image & Text                 &                       &                             &                                                                               &                                                                                & {}                       \\ 
\hline\hline
\multirow{12}{*}{\begin{tabular}[c]{@{}c@{}}MM-IMDB\\ (F1-Macro)\end{tabular}}    & \multirow{4}{*}{100\%} & \multirow{4}{*}{30\%}  & 100\% & 30\%                 & 32.78$\pm 0.29$         & 37.72$\pm 0.21$                      & 39.47$\pm 0.17$                                                                         & 40.01$\pm 0.19$                                                                          & \textbf{43.21}$\pm 0.19$                              \\
                                                                                  &                        &                        & 30\%  & 100\%                & 26.58$\pm 0.29$                 & 21.68$\pm 0.32$                       & 42.12$\pm 0.07$                                                                         & 49.72$\pm 0.17$                                                                          & \textbf{54.67}$\pm 0.14$                              \\
                                                                                  &                        &                        & 65\%  & 65\%                 & 30.55$\pm 0.17$                 & 30.80$\pm 0.59$                       & 40.99$\pm 0.06$                                                                         & 45.02$\pm 0.12$                                                                          & \textbf{49.09}$\pm 0.12$                              \\
                                                                                  &                        &                        & \multicolumn{2}{c|}{AVG}     & 29.97                 & 30.07                       & 40.86                                                                         & 44.91                                                                          & {\textbf{48.99}}         \\ 
\cline{2-10}
                                                                                  & \multirow{4}{*}{30\%}  & \multirow{4}{*}{100\%} & 100\% & 30\%                 & 30.25$\pm 0.30$                & 24.93$\pm 0.12$                       & 29.85$\pm 0.13$                                                                         & 39.60$\pm 0.25$                                                                          & \textbf{43.07}$\pm 0.31$                              \\
                                                                                  &                        &                        & 30\%  & 100\%                & 37.97$\pm 0.21$                 & 47.10$\pm 0.28$                     & 54.37$\pm 0.16$                                                                         & 50.94$\pm 0.26$                                                                          & \textbf{56.03}$\pm 0.14$                              \\
                                                                                  &                        &                        & 65\%  & 65\%                 & 34.45$\pm 0.44$                & 36.76$\pm 0.27$                      & 37.61$\pm 0.24$                                                                         & 45.83$\pm 0.15$                                                                          & \textbf{49.81}$\pm 0.16$                              \\
                                                                                  &                        &                        & \multicolumn{2}{c|}{AVG}     & 34.22                 & 36.26                       & 40.61                                                                         & 45.46                                                                          & {\textbf{49.64}}         \\ 
\cline{2-10}
                                                                                  & \multirow{4}{*}{65\%}  & \multirow{4}{*}{65\%}  & 100\% & 30\%                 & 35.80$\pm 0.29$                 & 39.04$\pm 0.15$                    & 40.60$\pm 0.26$                                                                         & 41.61$\pm 0.11$                                                                          & \textbf{42.46}$\pm 0.24$                              \\
                                                                                  &                        &                        & 30\%  & 100\%                & 36.65$\pm 0.16$                & 42.68$\pm 0.10$                      & 53.19$\pm 0.26$                                                                         & 52.05$\pm 0.14$                                                                          & \textbf{55.26}$\pm 0.22$                              \\
                                                                                  &                        &                        & 65\%  & 65\%                 & 36.66$\pm 0.27$                 & 41.33$\pm 0.48$                       & 47.34$\pm 0.24$                                                                         & 47.29$\pm 0.11$                                                                          & \textbf{49.24}$\pm 0.50$                              \\
                                                                                  &                        &                        & \multicolumn{2}{c|}{AVG}     & 36.37                 & 41.02                       & 47.04                                                                         & 46.98                                                                          & {\textbf{48.99}}         \\ 
\hline
\multirow{12}{*}{\begin{tabular}[c]{@{}c@{}}Food-101\\ (Accuracy)\end{tabular}}   & \multirow{4}{*}{100\%} & \multirow{4}{*}{30\%}  & 100\% & 30\%                 & 66.02$\pm 0.18$                 & 73.51$\pm 0.24$                       & 78.60$\pm 0.11$                                                                         & 78.50$\pm 0.18$                                                                          & \textbf{78.81}$\pm 0.13$                              \\
                                                                                  &                        &                        & 30\%  & 100\%                & 42.89$\pm 0.22$                 & 27.55$\pm 0.06$                       & 82.50$\pm 0.14$                                                                         & 86.03$\pm 0.14$                                                                          & \textbf{86.90}$\pm 0.10$                              \\
                                                                                  &                        &                        & 65\%  & 65\%                 & 54.49$\pm 0.12$                 & 50.59$\pm 0.16$                       & 80.37$\pm 0.22$                                                                         & 82.22$\pm 0.08$                                                                          & \textbf{82.35}$\pm 0.14$                              \\
                                                                                  &                        &                        & \multicolumn{2}{c|}{AVG}     & 54.47                 & 50.55                       & 80.49                                                                         & 82.25                                                                          & {\textbf{82.69}}         \\ 
\cline{2-10}
                                                                                  & \multirow{4}{*}{30\%}  & \multirow{4}{*}{100\%} & 100\% & 30\%                 & 42.80$\pm 0.31$                 & 29.61$\pm 0.16$                       & 66.50$\pm 0.20$                                                                         & 73.97$\pm 0.16$                                                                          & \textbf{75.41}$\pm 0.22$                              \\
                                                                                  &                        &                        & 30\%  & 100\%                & 76.54$\pm 0.12$                 & 86.45$\pm 0.11$                       & \textbf{87.44}$\pm 0.37$                                                                 & 87.17$\pm 0.11$                                                                           & 87.11$\pm 0.11$                                        \\
                                                                                  &                        &                        & 65\%  & 65\%                 & 59.54$\pm 0.19$                 & 57.97$\pm 0.15$                       & 76.98$\pm 0.10$                                                                          & 80.49$\pm 0.17$                                                                           & \textbf{81.22}$\pm 0.18$                               \\
                                                                                  &                        &                        & \multicolumn{2}{c|}{AVG}     & 59.63                 & 58.01                       & 76.97                                                                         & 80.55                                                                          & {\textbf{81.25}}         \\ 
\cline{2-10}
                                                                                  & \multirow{4}{*}{65\%}  & \multirow{4}{*}{65\%}  & 100\% & 30\%                 & 64.40$\pm 0.28$                 & 71.26$\pm 0.30$                       & 76.51$\pm 0.20$                                                                          & 77.67$\pm 0.16$                                                                           & \textbf{78.13}$\pm 0.15$                              \\
                                                                                  &                        &                        & 30\%  & 100\%                & 73.60$\pm 0.13$                 & 85.71$\pm 0.15$                       & 87.21$\pm 0.14$                                                                          & 86.79$\pm 0.12$                                                                           & \textbf{87.35}$\pm 0.12$                               \\
                                                                                  &                        &                        & 65\%  & 65\%                 & 68.96$\pm 0.27$                 & 78.43$\pm 0.16$                       & 81.82$\pm 0.14$                                                                          & 82.18$\pm 0.12$                                                                           & \textbf{82.67}$\pm 0.16$                               \\
                                                                                  &                        &                        & \multicolumn{2}{c|}{AVG}     & 68.99                 & 78.47                       & 81.85                                                                         & 82.21                                                                          & {\textbf{82.72}}         \\ 
\hline
\multirow{12}{*}{\begin{tabular}[c]{@{}c@{}}Hateful Memes\\ (AUROC)\end{tabular}} & \multirow{4}{*}{100\%} & \multirow{4}{*}{30\%}  & 100\% & 30\%                 & 60.77$\pm 0.72$                 & 58.54$\pm 1.27$                       & 59.77$\pm 0.63$& 57.65$\pm 0.50$& \textbf{61.39}$\pm 0.82$\\
                                                                                  &                        &                        & 30\%  & 100\%                & 61.84$\pm 0.42$                 & 55.67$\pm 1.30$                       & 66.41$\pm 0.28$& 67.36$\pm 0.25$& \textbf{66.50}$\pm 0.36$\\
                                                                                  &                        &                        & 65\%  & 65\%                 & 61.57$\pm 0.93$                 & 57.10$\pm 1.73$                       & 61.86$\pm 2.04$& 62.58$\pm 1.36$& \textbf{63.35}$\pm 1.18$\\
                                                                                  &                        &                        & \multicolumn{2}{c|}{AVG}     & 61.39                 & 57.10                       & 62.68                                                                         & 62.53                                                                          & {\textbf{63.75}}         \\ 
\cline{2-10}
                                                                                  & \multirow{4}{*}{30\%}  & \multirow{4}{*}{100\%} & 100\% & 30\%                 & 56.19$\pm 0.86$                 & 58.43$\pm 0.37$                       & 58.98$\pm 0.57$& 59.04$\pm 1.31$& \textbf{62.06}$\pm 0.38$\\
                                                                                  &                        &                        & 30\%  & 100\%                & 62.78$\pm 0.74$                 & 65.54$\pm 0.61$                       & 68.52$\pm 0.37$&\textbf{70.53}$\pm 0.10$& 68.97$\pm 0.30$\\
                                                                                  &                        &                        & 65\%  & 65\%                 & 59.29$\pm 1.02$                 & 62.36$\pm 0.35$                       & 63.58$\pm 0.16$& 64.78$\pm 0.98$& \textbf{65.12}$\pm 0.88$\\
                                                                                  &                        &                        & \multicolumn{2}{c|}{AVG}     & 59.42                 & 62.11                       & 63.69& 64.78                                                                          & {\textbf{65.38}}         \\ 
\cline{2-10}
                                                                                  & \multirow{4}{*}{65\%}  & \multirow{4}{*}{65\%}  & 100\% & 30\%                 & 59.42$\pm 0.92$                 & 60.05$\pm 1.40$                       & 60.93$\pm 0.46$& 60.39$\pm 0.27$& \textbf{61.39}$\pm 0.58$\\
                                                                                  &                        &                        & 30\%  & 100\%                & 63.13$\pm 0.45$                 & 61.88$\pm 1.07$                       & \textbf{69.19}$\pm 0.28$& 66.30$\pm 0.47$& 68.24$\pm 0.28$\\
                                                                                  &                        &                        & 65\%  & 65\%                 & 61.32$\pm 1.21$                & 61.52$\pm 1.11$                       & 64.47$\pm 0.63$& 62.90$\pm 0.47$& \textbf{65.17}$\pm 0.79$\\
                                                                                  &                        &                        & \multicolumn{2}{c|}{AVG}     & 61.29                 & 61.15                       & 64.86                                                                         & 63.20                                                                          & {\textbf{64.93}}       \\ 
\bottomrule
\end{tabular}}
\end{table*}
\clearpage
\subsection{Feature Prediction Loss Coefficients}

\begin{table}[]
\centering
\caption{Coefficient for feature prediction loss in Eq \ref{eq:vicreg} under a complete training setting. }
\setlength{\tabcolsep}{5pt} 
\renewcommand{\arraystretch}{1.2} 
\begin{tabular}{cccc}
\toprule 
$\lambda$ & $\mu$ & $\nu$ & F1-Macro \\ \hline \hline
5      & 5      & 1      &    36.11       \\
15     & 15     & 1      &    39.19       \\
25     & 25     & 1   &       38.42    \\
50     & 50     & 1   &  \textbf{40.88}     \\   \hline
\bottomrule 
\end{tabular}
\label{tab:my-table}
\end{table}

\end{document}